\documentclass[runningheads]{llncs}

\usepackage{times}
\usepackage{epsfig}
\usepackage{graphicx}
\usepackage{amsmath}
\usepackage{amssymb}
\usepackage{booktabs}
\usepackage{adjustbox}
\usepackage{multirow}
\usepackage{tabularx}
\usepackage[colorlinks]{hyperref}

\title{Recurrent Few-Shot model for Document Verification\thanks{This content reflects only the authors' view. The European Agency is not responsible for any use that may be made of the information it contains.}}
\titlerunning{Recurrent Few-Shot model for Document Verification}
\author{Maxime Talarmain \orcidID{0009-0005-2168-2025}  \and
Carlos Boned  \orcidID{0009-0000-6041-0931} \and
Sanket Biswas  \orcidID{0000-0001-6648-8270} \and 
Oriol Ramos Terrades\orcidID{0000-0002-3333-8812}
 }
\authorrunning{M. Talarmain et al.}
\institute{Computer Vision Center \and
Computer Science Department \\
Universitat Autònoma de Barcelona, Catalunya \\
\email{\{mtalarmain, cboned, sbiswas, oriolrt\}@cvc.uab.cat} }

\begin{document}


\maketitle
\thispagestyle{empty}

\begin{abstract}
   General-purpose ID, or travel, document image- and video-based verification systems have yet to achieve good enough performance to be considered a solved problem. There are several factors that negatively impact their performance, including low-resolution images and videos and a lack of sufficient data to train the models. This task is particularly challenging when dealing with unseen class of ID, or travel, documents. In this paper we address this task by proposing a recurrent-based model able to detect forged documents in a few-shot scenario. The recurrent architecture makes the model robust to document resolution variability. Moreover, the few-shot approach allow the model to perform well even for unseen class of documents. Preliminary results on the SIDTD and Findit datasets show good performance of this model for this task.
\end{abstract}

\section{Introduction}\label{sec:intro}
The increased of remote identity authentication systems, incorporating biometrics and the verification of ID and travel documents, has surged and become widespread in the wake of the COVID-19 pandemic. These authentication systems have empowered citizens to engage in work and business activities outside traditional office settings. Public administration, banks, productive industries, and numerous services have integrated these systems seamlessly into their routine workflows. These services provide an online enrollment option, eliminating the need for users to physically attend by requesting a selfie and an image of their ID document for authentication. Nevertheless, cybercrime has exploited societal vulnerabilities, evolving towards increasingly sophisticated threats. Therefore, it is necessary to develop techniques that allow us to detect this type of fraudulent actions that expose citizens' data and their privacy to the general public.

A current trend is to detect ID documents, passports or driving licenses that have physically or digitally been modified from images acquired from mobile devices. 
In \cite{Berenguel2017}, the authors applied texture descriptors to image patches and then applied BoW followed by an SVM classifier to classify each patch as genuine or fake. In that work, the most performant CNN architectures of that time (AlexNet, VGG and Inception)  were also used as general descriptor extractors. The results obtained in terms of F1-Score were very good, obtaining results between 0.99 and 1 in many cases. The main problem of that kind of approach was the need for more sufficient examples of forged documents and the poor generalization capacity for unseen classes of documents during the training process.

The results in \cite{Berenguel2017} are consistent with recent works using more advanced deep learning techniques. In \cite{Saire2020}, Residual networks are applied to detect forged documents at different scales. In \cite{Yan2021}, the authors also apply Residual networks to study the impact of using different scanners and printers to detect images of genuine documents compared to recaptured documents. On the other hand, the authors in \cite{Chen2021} propose the use of Siamese networks for the detection of the recapture of ID documents (student ID Cards in their experiments). Encoder-Decoder architectures with attention modules have also been used to address this problem in \cite{Liang2022}. Recently, spatial and spectral features have been used for the localization of modified regions~\cite{CHEN2022_PR}. In all these works, although the reported results show good yields (above 0.90 in AUC and similar metrics), with a certain capacity for generalization depending on the recapture devices used. When they have to be applied to different types of documents, like for instance ID cards from nationalities not seen in training, the performance of these methods (and similar ones) methods drops dramatically~\cite{Berenguel2019}. 

Consequently, and due to the inability to obtain sufficient representative learning data, both in terms of enough samples of forged documents and the paramount variability of genuine data, the transfer of this technology to products working in real scenarios is still far from being a reality. The results reported in the  above-mentioned works, and others, show that the problem is not in the discrimination power  of current models  but of the incapacity of models  to generalize to a real environment, where the indefinability of the types of documents and the unpredictability of the type of forgeries is the common scenario. 

In this paper, we go a step beyond the generalization capacity of document verification models. We combine pre-trained models with a recurring network in a very simple way. The weights of this network are trained in a few-shot learning context. The results obtained are surprisingly good, even when we apply the models trained with the ID Documents dataset to a ticket images dataset. The achieved performance of the proposed model opens new perspectives for this technology.

\section{Methodology}
\label{sec:method}

\begin{figure}[t]
    \centering
    \includegraphics[width=\textwidth]{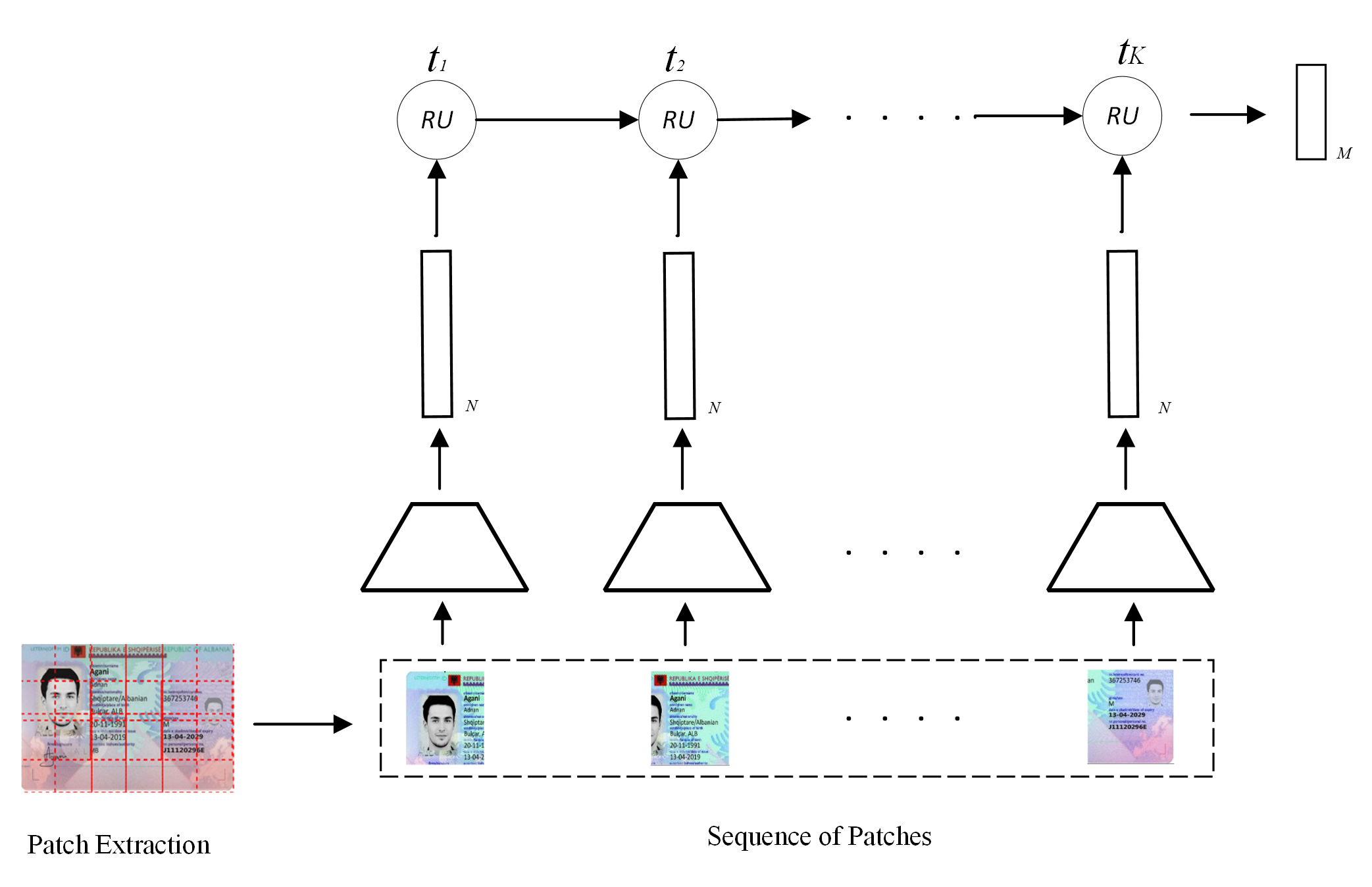}
    \caption{Data modelling architecture: The sequence of patches are extracted for each document. Each patch is then fed into a CNN feature extractor. The vectors from the CNN are passed to a RU at time $t_k$ depending of the patch position. The overall vector of the document is extracted by the RU at $t_K$ }
    \label{fig:model_archi}
\end{figure}

As we mentioned in the Introduction, the proposed model is relatively simple and, to some extent, unoriginal as it combines components that are well known in the community. \figurename~\ref{fig:model_archi} shows  the architecture used to train the model. Essentially, it is a recurring many-to-one network. The input of each recurring unit, in addition to the previous state of the network, is a feature vector calculated in patches of the image. The output of the network is  an $m$-dimensional vector  that is used to classify the document image  as genuine or fake. 

Document images have been partitioned into $W$-dimensional square patches that overlap  to avoid contour effects. Each patch is processed by a pre-trained backbone that is used as an universal feature extractor. The output of the backbone is a vector of dimension $n > m$ that feeds the Recurrent Unit (RU).

We apply the model in \figurename~\ref{fig:model_archi} to both the support set and the query set to apply a few-shot learning (FSL) strategy to classify the documents based on their vector representation. FSL training is structured around the concept of episodes. At each episode, the model is learning from a support set composed of $k$-shots and $n$-ways, and then, after $N_1$ episodes where $N_1<N$, we do $M$ episodes of evaluation on the query set with $q$-shots and $n$-ways. $n$-ways here represents the number of class labels present in the support or query set at each episode, therefore $n$ will always be equal to 2. The parameter $k$-shots (eq. $q$-shots) represents the number of images allocated to each class label within the support set (eq. query set) during every episode. Hence, at each episode on support set, the model is learning from $n \times k$ images, and evaluating on $n \times q$ images on query set.
At each episode, the batch of images are grouped by class label, and the order of class label is shuffled at each episode, which means, the first $k$ images (or $q$ images for query set) could be either forged or genuine depending on the episode.
It is worth to mention that support set and query set do not contain the same meta-class, as the main objective is to build a model able to generalize on unseen meta-classes. In this work, we defined meta-classes based on the country of issuance of the ID document.

From the support set vectors and the query set vectors, we learn a metric space based on Prototypical Network (PN) method, that will classify  document images in the corresponding class, either genuine or faked. The loss used during training is a cross-entropy loss. We trained the whole architecture using two distinct strategies: Conditional FSL and Unconditional FSL. The Conditional FSL assumes that the system knows beforehand the document class (the meta-class in a few-shot scenario). Conversely, the Unconditional FSL does not knows the document class and must detect it is genuine or not. 

\subsection*{C-FSL: Conditional few-shot learning}

\begin{figure}[t]
    \centering
    \includegraphics[width=\textwidth]{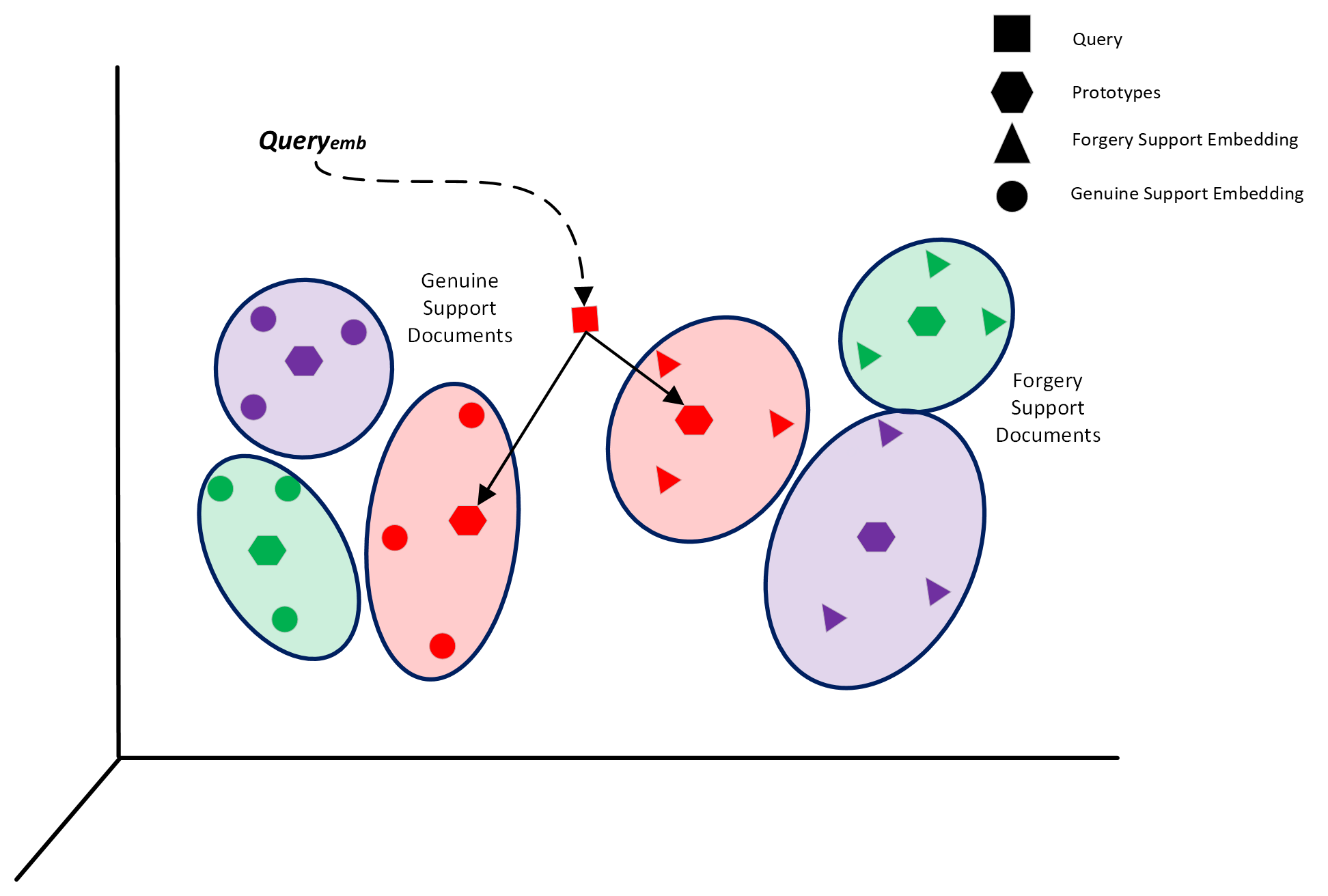}
    \caption{Example of conditioned prototypical classification with k-shot=3}
    \label{fig:proto_classif}
\end{figure}

This approach involves training the model on a support set and a query set derived from the same document meta-class. As depicted in \figurename~\ref{fig:proto_classif}, different meta-classes are distinguished by distinct colors.

The final prediction for a query vector is given by its closest distance to the support set vectors within the same meta-class. By adopting this  approach, the model is relieved from the burden of accommodating various document types in the  space, allowing it to concentrate solely on detecting fake documents, see \figurename~\ref{fig:proto_classif}. 

In this scenario, as shown in the figure, triangles represent fake documents, while circles represent genuine documents. Each pentagon represents the mean vector for each meta-class, and the square represents  the query vector computed from the recurrent network model. The learnt distance is computed from the query vector to the support vectors belonging to the same meta-class. If the closest meta-class  distance belongs to support vectors of fake documents, the query is classified as a fake document.

\subsection*{U-FSL: Unconditional few-shot learning}

\begin{figure}[t]
    \centering
    \includegraphics[width=\textwidth]{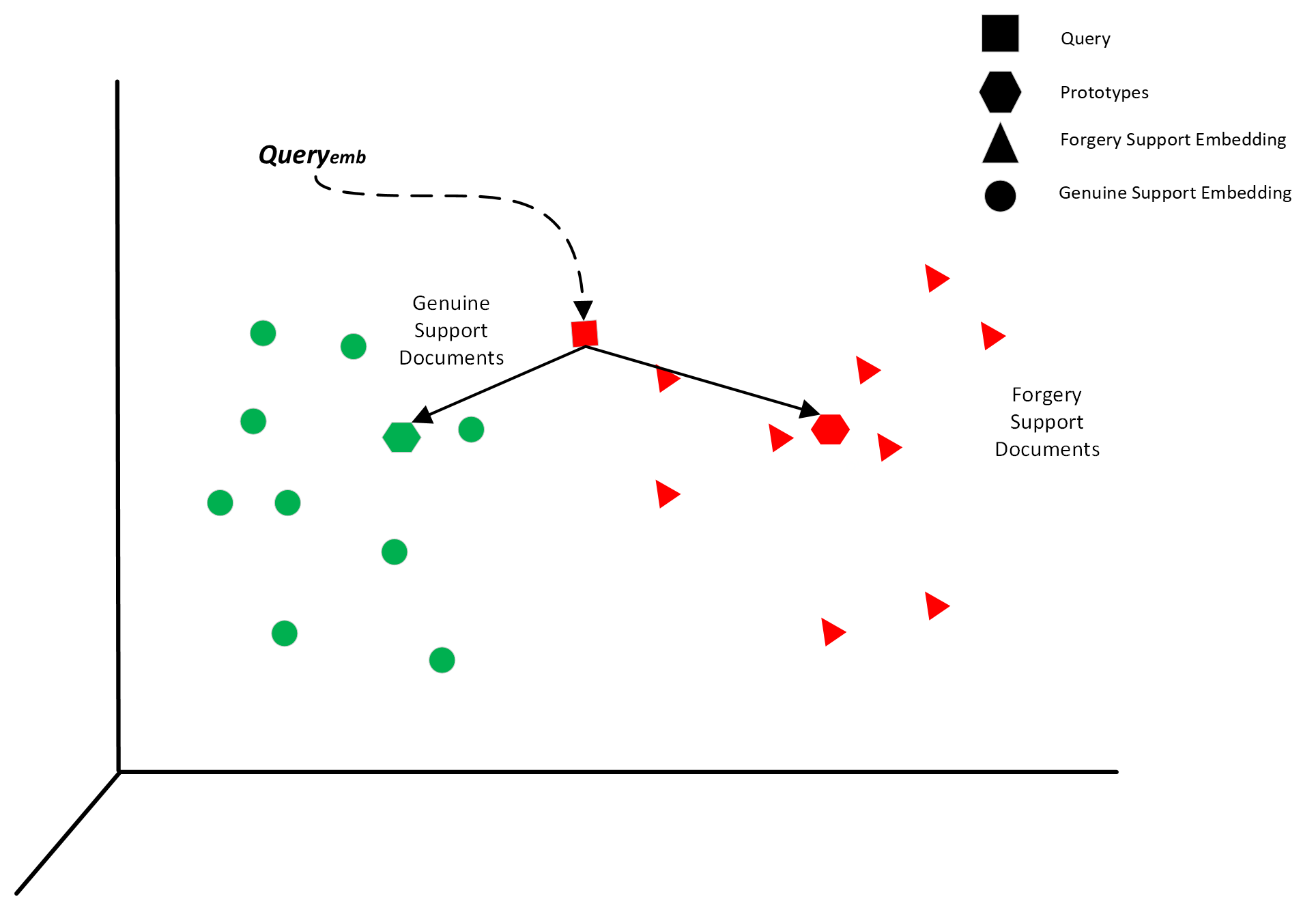}
    \caption{Example of Unconditioned Prototypical classification with k-shot=10}
    \label{fig:uncon_classif}
\end{figure}

As explained above, in the Unconditional few-shot strategy, all meta-classes are included in both the support and query sets without considering the country of origin. This ensures that each support  vector exclusively represents either genuine or fake documents, regardless of the document meta class. This approach requires models to focus solely on identifying fake document  and inconsistencies, while also minimizing the impact of meta-class variations in document representations. \figurename~\ref{fig:uncon_classif} illustrates this concept, depicting data distribution without conditioning on meta-classes, where both genuine and fake classes remain. In this case the query will be classified as genuine as the distance to the genuine prototype is closer as the distance to the support vectors of fake documents.

\section{Related Work}\label{sec:sota}

Document verification is essentially a binary classification task. We have seen in the Introduction that in a classical supervised context almost any classification method would perform well if it were not for the lack of sufficient quality data to provide these models enough generalization capacity. In this section, we will review the key architecture components that closely related to the proposed model. Therefore, we will briefly review the main convolution architectures that can be used as backbones in our model, the usual recurring units, and the main FSL strategies. 

The proposed architecture, illustrated in \figurename~\ref{fig:model_archi}, applies a {\em convolutional} module as an universal feature extractor. Although there are many models  that can be used as backbone and are trained on large databases such as ImageNet, we use EfficientNet-B3~\cite{tan2019efficientnet}, ResNet50 \cite{he2016deep}, Vision Transformer Small Patch 16 (ViT-S/16)~\cite{dosovitskiy2020vit} and TransFG~\cite{he2022transfg}, since we want to evaluate the impact to these pre-trained models in the final performance of the proposed model. EfficientNet and ResNet belong to the commonly used architectures for image classification tasks while Vision Transformer and TransFG architectures are transformer-based architectures. Moreover, the TransFG model is an architecture designed for fine-grained classification task.

Although we have used attention-based models as a backbone, the main element of the proposed architecture is the Recurrent Unit (RU) that we apply to the output of them. As the patches order within the sequence  is completely arbitrary we perform some preliminary experiments to evaluate the impact  in the model performance, depending on the patches order. The impact on the model performance was negligible and thus, a natural choice would be to apply Attention-based models~\cite{vaswani2017attention}. However, since recurrent models perform  equally, we choose  simpler models that require less data to train. The RUs used and evaluated are the usual RNN, LSTM~\cite{hochreiter1997long} and GRU~\cite{cho2014learning}.

To address the problem of having to estimate models able of generalizing with a limited set of data, the authors in ~\cite{snell2017prototypical} developed the method Prototypical Networks. In this work, prototypes are computed from a reduced set of samples per class and  are used to estimate a metric space that allows samples classification based on the square Euclidean distance between prototypes and the query. In ~\cite{vinyals2016matching}, the authors proposed a paradigm called Matching Networks (MN), which used a differentiable nearest neighbors algorithm for FSL classification. Also, the authors in ~\cite{finn2017model} introduced the Model-Agnostic Meta-Learning (MAML) technique, a strategy that updates model parameters with a small number of gradient steps and limited train data points from a new task to produce a good generalization performance. These approaches have shown promise in domains, including image recognition and natural language processing, yet their application in document ID verification remains underexplored.

\section{Experiments and discussion}\label{sec:Experiments}

As we have pointed from the beginning of this paper, the biggest weakness of document verification methods in general is their unreliability when they have to process document classes not seen in training. Consequently, the goal of the experiments carried out was to evaluate as much as possible the generalization capacity  of the proposed model. To this end, we have repeated 10 times the same experiments with randomly chosen metaclasses,  queries sets and support sets. We reported the performances of the evaluated models in terms of the  the accuracy rate  and the area under the curve, respectively denoted by Accuracy and AUC in the results tables. Together with the mean of the values of these metrics, we have computed their standard deviation. Below, we briefly describe the datasets used and the experiments set up. We conducted an ablation study to assess the impact of the backbone and the recurrent units in the model performance. Then, we analyse and discuss the generalization capacity of the proposed models under harder conditions. 

\begin{figure}[!t]
    \centering
    \begin{tabular}{cc}
        \includegraphics[width=.5\textwidth]{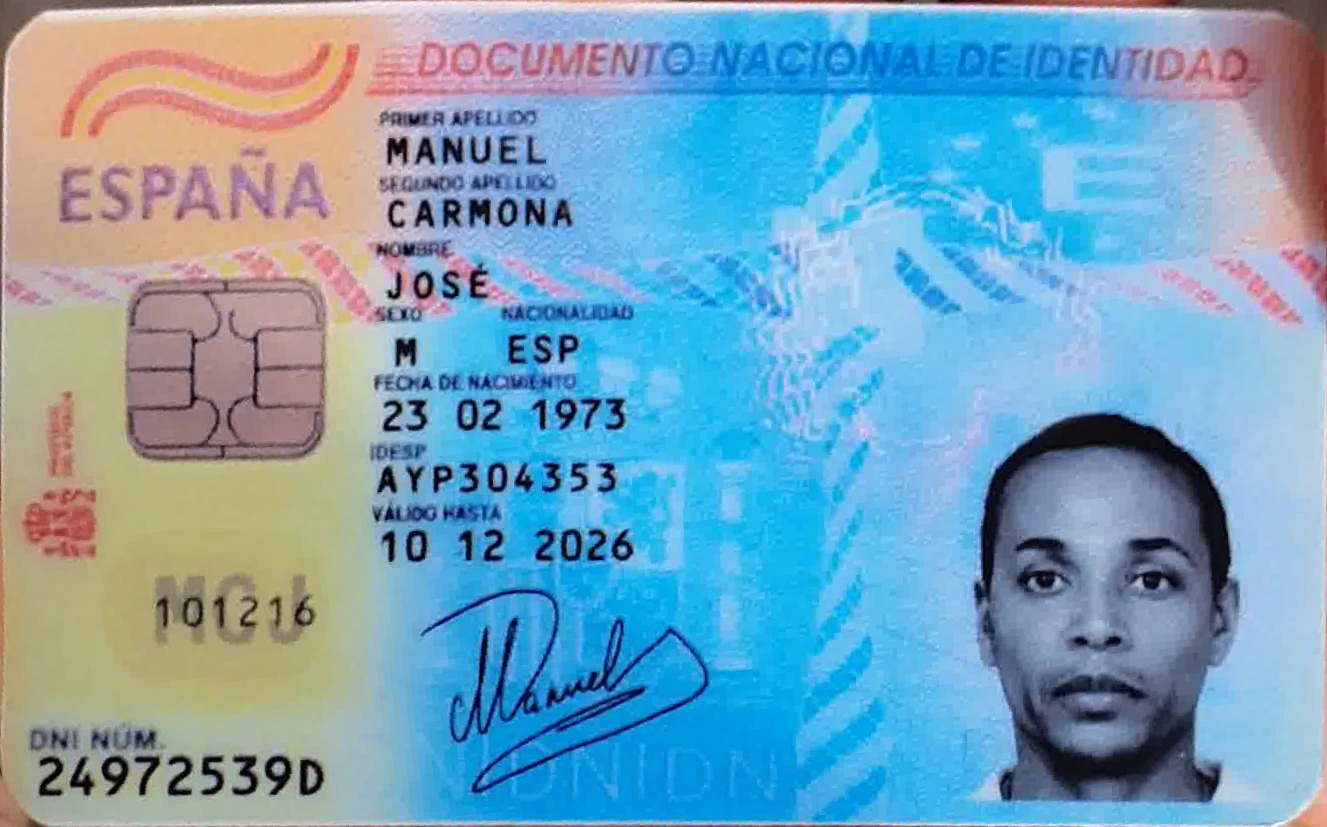} &
            \includegraphics[width=.5\textwidth]{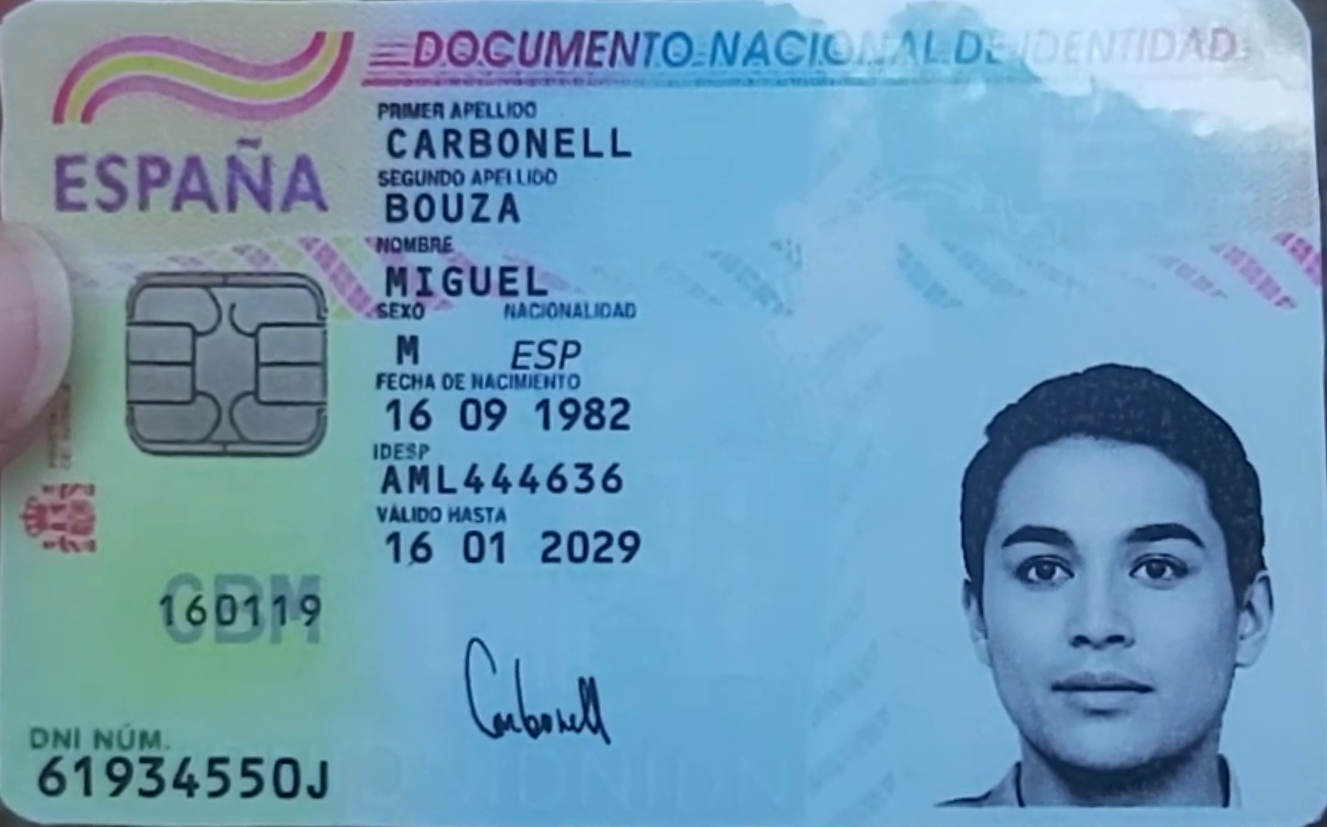} \\
            (a) Genuine example & (b) Fake example
    \end{tabular}
    \caption{Example of cropped images from the SIDTD clips. (a) corresponds to a genuine sample of a synthetic Spanish ID document and (b) to a fake sample of the same nationality.}
    \label{fig:SIDTD_example}
\end{figure}

\subsection*{Datasets}

\begin{figure}[!t]
    \centering
    \includegraphics[width=\textwidth]{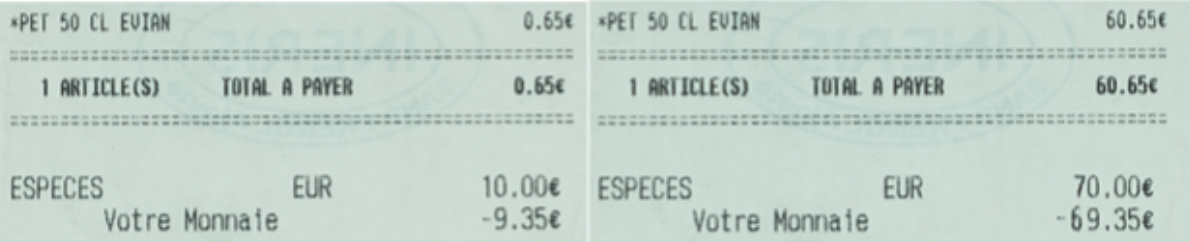}
    \caption{Example of forgeries on prices, left image is a genuine receipt and the right image is altered from the FindIT dataset.}
    \label{fig:findit_ex}
\end{figure}

We trained models on the SIDTD dataset~\cite{boned2024synthetic}. The SIDTD dataset is an extension of the MIDV2020 dataset\cite{Bulatov_2022} and it contains ID cards in three different formats: templates, videos, and clips. Each format includes both genuine and fake documents. For the evaluation purposes we used the clips images, which are the video frames, and cropped to remove image background and to rectify the perspective effects, see \figurename~\ref{fig:SIDTD_example}. In addition to the SIDTD dataset we use two more datasets for testing purposes. The first dataset, named in this paper as {\em UAB dataset},  includes real Spanish National ID Cards voluntarily provided by people. To generate forgeries of these documents and create fake examples, we employed the same techniques used in generating forgeries for the SIDTD dataset. The second dataset used for testing purposes was the Findit dataset, which is another fraud detection dataset based on French receipts~\cite{artaud2018find}. The Findit dataset provides 1,180 images of captured receipts (240 altered and 940 genuine receipts) modified by real people from common and widespread material such as Window 10® to obtain closer real-life forgery, see \figurename~\ref{fig:findit_ex}.

\subsection*{Experimental Set up}
For training, and following the FSL methodology\cite{Wang2020}, we have divided the meta-train and meta-test based on the origin of the document. The training set is composed of 6 document nationalities, and in the test set there are the other 4 nationalities. We iterated ten times the same training procedures, with distinct document nationalities for both the training and testing sets, by randomly permuting the meta-classes at each iteration. Each document nationality  has an average of 100-140 images.

Images are initially resized to the average shape found in the SIDTD dataset, which  is $1047\times1564$. Then, we have set the patch size  299 width, generating 70 patches per image. The batch composition for training varies based on the chosen strategy. For the C-FSL strategy, we adopted a 5-shots approach, wherein 5 fakes and 5 genuine samples  serve as support, and another set of 5 fakes and 5 genuine samples act as queries. In the U-FSL strategy, we defined a 10-shots approach with the same distribution of fakes and genuine samples as described before.

Training was conducted on an NVIDIA A40 with 40GB of memory for 5,000 episodes and using the Adam optimizer. For the backbone, we trained on diverse standard CNN Backbones, including EfficientNet-B3 \cite{tan2019efficientnet}, ResNet50 \cite{he2016deep}, Vision Transformer Small Patch 16 (ViT-S/16) \cite{dosovitskiy2020vit} and TransFG \cite{he2022transfg}, all pretrained on ImageNet. The recurrent cell was trained on three standard recurrent cells: RNN, LSTM,  and GRU.

\subsection*{Discussion}
\label{sec:discussion}

The results obtained are quite good overall. If we simply compare the performance metrics with the recently published ones, it can give the impression that there is not that much difference. However,  remember that in the results reported in \cite{Chen2021,CHEN2022_PR,Liang2022}, among others, the experimental framework corresponds to the usual supervised learning one. On the other hand, our experimental framework corresponds to a FSL framework.  In this sense, the results reported in \cite{Berenguel2019} achieved an AUC close to 0.90 in the most restrictive scenario with the best of the reported methods. In this paper we obtain AUC above 0.95 in the worst-case scenario in all cases.

Experiments comparing the impact of the RU on overall performance have shown us that, surprisingly, LSTMs perform better than GRUs, when this is not the norm in other contexts. Likewise, in relation to the FSL strategies, it has been with the simplest one, the PN, that the best results have been obtained. In what follows, we will discuss the results obtained by evaluating both U-FSL and C-FSL models for each of the backbones used.

\tablename~\ref{tab:all_results_backbone} provides a comprehensive overview of the results obtained for each backbone. There are some evidences that, for both C-FSL and U-FSL strategies, the best results are consistently achieved when using ResNet as the backbone. Again, simplest models achieves the best performance. Additionally, we observe that the results of the C-FSL strategy are slightly better to those of the U-FSL strategy. This is not surprising since the conditional model, by using the information from the metaclass, reduces the uncertainty of the model and significantly simplifies the task. What is surprising is that the decay in performance, looking at the standard deviation values of the experiments, does not seem significant for the U-FSL strategy. 

\begin{table}[!t]
\centering
\begin{tabular}{|c|c|c|c|c|} 
\hline
\multicolumn{5}{|c|}{Clip cropped SIDTD}\\
\hline
 \multirow{2}{*}{Models}  & \multicolumn{2}{c|}{U-FSL}  &   \multicolumn{2}{c|}{ C-FSL} \\   \cline{2-5} 
  & Accuracy & AUC & Accuracy & AUC \\
\hline 
EffNet & $98.50 \pm 0.97$ & $99.95 \pm 0.05$  & $99.34 \pm 0.47$ & $99.98 \pm 0.03$  \\  
ResNet & $99.65 \pm 0.25$ & $100 \pm 0.0$ & $99.98 \pm 0.07$ & $100 \pm 0.0$  \\ 
\hline 
ViT & $98.48 \pm 0.08$ & $99.90 \pm 0.01$ & $99.91 \pm 0.11$ & $100 \pm 0.0$    \\  
TransFG & $98.20 \pm 0.46$ & $99.92 \pm 0.04$ &  $99.89 \pm 0.19$ & $100 \pm 0.0$ \\ 
\hline 
\end{tabular}
\caption{PN and LSTM model performances in  terms of Accuracy and AUC  scores with standard deviation. Models are trained and tested on the clip cropped SIDTD images. Training is performed with 5-shot for C-FSL models and with 10-shot with U-FSL models. }
\label{tab:all_results_backbone}
\end{table}

Previous experiments have been conducted by scaling the images to the average size of the images. However, one of the main benefits of recurrent networks is we can process sequences of different lengths. In this case, we can process document images of different sizes (or resolution). In the following experiment, with the models trained in the previous experiments, we have evaluated the robustness of our model when processing images of documents of different resolution. To that end, the images in the SIDTD clip partition have been cropped and rectified but preserving the scale. The dimensions of the processed images range from 211 pixels wide (the size of a patch) to 4,228 pixels. \tablename~\ref{tab:no scale} presents the results obtained and are equally satisfactory for both, the C-FSL  model and the U-FSL model. Again, the C-FSL model achieves consistently  better performance than U-FSL models but still, the performance of the U-FSL model are quite good. What deserves further analysis is the standard deviation values, since overall it has increase with respect the first set of experiments. We must analyse whether there is a correlation between the performance and the document resolution, or not, which will make sense.

\begin{table}[!t]
\centering
\begin{tabular}{|c|c|c|c|c|} 
\hline
\multicolumn{5}{|c|}{clip cropped SIDTD No Rescaled}\\
\hline
 \multirow{2}{*}{Models}  & \multicolumn{2}{c|}{U-FSL}  &   \multicolumn{2}{c|}{C-FSL} \\   \cline{2-5} 
  & Accuracy & AUC & Accuracy & AUC \\
\hline 
EffNet & $95.21 \pm 3.67$ & $99.13 \pm 0.93$ & $97.18 \pm 6.03$ & $99.53 \pm 1.14$  \\  
ResNet & $97.17 \pm 2.79$ & $99.79 \pm 0.31$ & $99.99 \pm 0.01$ & $100 \pm 0.0$ \\ 
\hline 
ViT & $92.96 \pm 4.40$ & $98.31 \pm 1.16$ & $99.92 \pm 0.12$ & $100 \pm 0.0$  \\  
TransFG & $95.87 \pm 0.29$ & $99.44 \pm 0.50$ & $96.75 \pm 4.01$ & $99.68 \pm 0.37$ \\ 
\hline 
\end{tabular}
\caption{PN and LSTM  model performances in  terms of Accuracy and AUC  scores with standard deviation. Models are tested on clip cropped SIDTD images not  rescaled. Inference is performed with 5-shot for the  C-FSL model and with 10-shot for the U-FSL model. }
\label{tab:no scale}
\end{table}

The last set of experiments seeks to evaluate the generalization capacity of the proposed models in a more challenging scenario. The models trained in the first experiment have been evaluated on two different datasets. As we have previously described, the UAB dataset is made up of real ID documents that have been forged with the same techniques used to generate the fake data of the SIDTD dataset. The second dataset has nothing to do with the type of documents that have been used to train the models and is made up of images of tickets that have been altered by different means. For the latter dataset, since there is only one metaclass, the U-FSL and C-FSL models are the same. Results  are reported in \tablename~\ref{tab:all_results_uab_dataset}. Again the best performance is achieved with the ResNet as backbone. Moreover, the performance achieved in terms of Accuracy and AUC for both datasets is aligned with the performance achieved in previous experiments.

\begin{table}[!t]
\centering
\begin{tabular}{|c|c|c|c|c|c|c|} 
\hline
Dataset & \multicolumn{4}{c|}{UAB Dataset} & \multicolumn{2}{c|}{Findit}\\
\hline
 \multirow{2}{*}{Models}  & \multicolumn{2}{c|}{U-FSL}  &   \multicolumn{2}{c|}{C-FSL} & \multicolumn{2}{c|}{U-FSL} \\   \cline{2-7} 
  & Accuracy & AUC & Accuracy & AUC & Accuracy & AUC \\
\hline 
EffNet & $96.09 \pm 1.69$ & $99.65 \pm 0.27$  & $99.98 \pm 0.02$ & $100 \pm 0.0$  & $89.93 \pm 7.83$ & $94.60 \pm 7.78$  \\  
ResNet & $98.77 \pm 1.49$ & $99.39 \pm 0.10$ & $100 \pm 0.0$ & $100 \pm 0.0$ & $98.56 \pm 1.69$ & $99.93 \pm 0.10$ \\ 
\hline 
ViT & $90.08 \pm 7.17$ & $95.85 \pm 6.57$ & $99.62\pm 0.60$ & $99.99 \pm 0.03$  & $89.87 \pm 8.99$ & $95.26 \pm 8.43$   \\  
TransFG & $97.65 \pm 0.82$ & $99.83 \pm 0.09$ &  $99.79 \pm 0.24$ & $99.99 \pm 0.01$ & $96.50 \pm 1.54$ & $99.65 \pm 0.27$ \\ 
\hline 
\end{tabular}
\caption{PN and LSTM model performances in  terms of Accuracy and AUC  scores with standard deviation. Models are tested on UAB Dataset and FindIt. Inference is performed with 5-shot for C-FSL models and with 10-shot for U-FSL models. }
\label{tab:all_results_uab_dataset}
\end{table}

As we mentioned at the beginning of this subsection, overall the results obtained are quite good. The ablation study performed by varying  backbones, recurrent cells and FSL strategies shows that there are no big differences between the use of some components with respect to others. However, we observed that simpler components tend to perform better than more complex ones. In particular, the relative poor performance of pre-trained attention-based backbones (ViT and TransFG) is striking, when the general trend in other areas of application is just the opposite. Finally, the idea of making use of a recurring network along with a FSL strategy seems to have been successful. The combination of both techniques in meta-training has resulted in more robust representation that allows us to generalize well to the proposed model.

\section{Conclusion}\label{sec:conclusion}

In this paper, we have presented a recurrent network model combined with FSL strategies to verify whether document images are genuine or fake. Despite of not introducing any new component in this architecture, the proposed architecture, by itself, is original and has not been applied to this task and similars. Moreover, the results obtained, as we have already discussed in the previous section, support the proposed strategy.

The proposed model seems to be able to learn good document representations. That representations are what would allow our proposed model to generalize well. However, there are still elements that deserve further study. FSL models still require few examples of fake and genuine documents and it is knows that in practice is not always feasible. Therefore, it is necessary to develop models that allow us to move towards zero-shot models.

\section{Acknowledgements}
This work has been partially supported by the SOTERIA project, which has received funding from the European Union’s Horizon 2020 research and innovation program under grant agreement No 101018342. It is also partially supported by the Spanish project PID2021-126808OB-I00 (GRAIL), Ministerio de Ciencia e Innovación,  CNS2022-135947 (DOLORES) and the AGAUR SGR project 2021-SGR-01559.  The authors acknowledge the support of the Generalitat de Catalunya CERCA Program to CVC’s general activities.

{\small
\bibliographystyle{ieee_fullname}
\bibliography{main}

\begin{thebibliography}{10}\itemsep=-1pt

\bibitem{artaud2018find}
C. Artaud, N. Sidere, A. Doucet, J.-M. Ogier, and V. Poulain~D'Andecy.
\newblock Find it! fraud detection contest report.
\newblock In {\em 2018 24th International Conference on Pattern Recognition (ICPR)}, pages 13--18, 2018.

\bibitem{boned2024synthetic}
C. Boned, M. Talarmain, N. Ghanmi, G. Chiron, S. Biswas, A.~M. Awal, and O.~Ramos Terrades.
\newblock Synthetic dataset of id and travel document, 2024.

\bibitem{Bulatov_2022}
K.B. Bulatov, E.V. Emelianova, D.V. Tropin, N.S. Skoryukina, Y.S. Chernyshova, A.V. Sheshkus, S.A. Usilin, Z. Ming, J.-C. Burie, M.M. Luqman, and V.V. Arlazarov.
\newblock Midv-2020: a comprehensive benchmark dataset for identity document analysis.
\newblock {\em Computer Optics}, 46(2), Apr. 2022.

\bibitem{Berenguel2017}
A.~Berenguel Centeno, O.~Ramos Terrades, J.~Llad{\'{o}}s Canet, and C.~Ca{\~{n}}ero Morales.
\newblock Evaluation of texture descriptors for validation of counterfeit documents.
\newblock In {\em 14th {IAPR} International Conference on Document Analysis and Recognition, {ICDAR} 2017, Kyoto, Japan, November 9-15, 2017}, pages 1237--1242, 2017.

\bibitem{Berenguel2019}
A.~Berenguel Centeno, O.~Ramos Terrades, J.~Llad{\'{o}}s Canet, and C.~Ca{\~{n}}ero Morales.
\newblock Recurrent comparator with attention models to detect counterfeit documents.
\newblock In {\em 2019 International Conference on Document Analysis and Recognition, {ICDAR} 2019, Sydney, Australia, September 20-25, 2019}, pages 1332--1337. {IEEE}, 2019.

\bibitem{Chen2021}
C. Chen, S. Zhang, F. Lan, and J. Huang.
\newblock Domain-agnostic document authentication against practical recapturing attacks.
\newblock {\em IEEE Transactions on Information Forensics and Security}, 17:2890--2905, 2022.

\bibitem{CHEN2022_PR}
C. Chen, L. Zhao, J. Yan, and H. Li.
\newblock A distortion model-based pre-screening method for document image tampering localization under recapturing attack.
\newblock {\em Signal Processing}, 200:108666, 2022.

\bibitem{cho2014learning}
K. Cho, B. Van~Merri{\"e}nboer, C. Gulcehre, D. Bahdanau, F. Bougares, H. Schwenk, and Y. Bengio.
\newblock Learning phrase representations using rnn encoder-decoder for statistical machine translation.
\newblock {\em arXiv preprint arXiv:1406.1078}, 2014.

\bibitem{dosovitskiy2020vit}
A. Dosovitskiy, L. Beyer, A. Kolesnikov, D. Weissenborn, X. Zhai, T. Unterthiner, M. Dehghani, M. Minderer, G. Heigold, S. Gelly, et~al.
\newblock An image is worth 16x16 words: Transformers for image recognition at scale.
\newblock {\em arXiv preprint arXiv:2010.11929}, 2020.

\bibitem{finn2017model}
C. Finn, P. Abbeel, and S. Levine.
\newblock Model-agnostic meta-learning for fast adaptation of deep networks.
\newblock In {\em International conference on machine learning}, pages 1126--1135. PMLR, 2017.

\bibitem{he2022transfg}
J. He, J.~N. Chen, S. Liu, A. Kortylewski, C. Yang, Y. Bai, and C. Wang.
\newblock Trans{FG}: A transformer architecture for fine-grained recognition.
\newblock In {\em Proceedings of the AAAI Conference on Artificial Intelligence}, volume~36, pages 852--860, 2022.

\bibitem{he2016deep}
K. He, X. Zhang, S. Ren, and J. Sun.
\newblock Deep residual learning for image recognition.
\newblock In {\em Proceedings of the IEEE conference on computer vision and pattern recognition}, pages 770--778, 2016.

\bibitem{hochreiter1997long}
S. Hochreiter and J. Schmidhuber.
\newblock {L}ong {S}hort-{T}erm {M}emory.
\newblock {\em Neural computation}, 9(8):1735--1780, 1997.

\bibitem{Liang2022}
W. Liang, L. Dong, R. Wang, D. Yan, and Y. Li.
\newblock Robust document image forgery localization against image blending.
\newblock In {\em 2022 IEEE International Conference on Trust, Security and Privacy in Computing and Communications (TrustCom)}, pages 810--817, 2022.

\bibitem{Saire2020}
D. Saire and S. Tabbone.
\newblock Documents counterfeit detection through a deep learning approach.
\newblock In {\em 2020 25th International Conference on Pattern Recognition (ICPR)}, pages 3915--3922, 2021.

\bibitem{snell2017prototypical}
J. Snell, K. Swersky, and R.~S. Zemel.
\newblock Prototypical networks for few-shot learning, 2017.

\bibitem{tan2019efficientnet}
M. Tan and Q. Le.
\newblock {Efficientnet}: {R}ethinking model scaling for convolutional neural networks.
\newblock In {\em International conference on machine learning}, pages 6105--6114, 2019.

\bibitem{vaswani2017attention}
Ashish Vaswani, Noam Shazeer, Niki Parmar, Jakob Uszkoreit, Llion Jones, Aidan~N Gomez, {\L}ukasz Kaiser, and Illia Polosukhin.
\newblock Attention is all you need.
\newblock {\em Advances in neural information processing systems}, 30, 2017.

\bibitem{vinyals2016matching}
O. Vinyals, C. Blundell, T. Lillicrap, D. Wierstra, et~al.
\newblock Matching networks for one shot learning.
\newblock {\em Advances in neural information processing systems}, 29, 2016.

\bibitem{Wang2020}
Y. Wang, Q. Yao, J.~T. Kwok, and L.~M. Ni.
\newblock Generalizing from a few examples: A survey on few-shot learning.
\newblock {\em ACM Comput. Surv.}, 53(3), jun 2020.

\bibitem{Yan2021}
J. Yan and C. Chen.
\newblock Cross-domain recaptured document detection with texture and reflectance characteristics.
\newblock In {\em 2021 Asia-Pacific Signal and Information Processing Association Annual Summit and Conference (APSIPA ASC)}, pages 1708--1715, 2021.

\end{thebibliography}
}

\end{document}